\documentclass[sigconf]{acmart}
\copyrightyear{2024}
\acmYear{2024}
\setcopyright{rightsretained}
\acmConference[KDD '24]{Proceedings of the 30th ACM SIGKDD Conference on Knowledge Discovery and Data Mining}{August 25--29, 2024}{Barcelona, Spain}
\acmBooktitle{Proceedings of the 30th ACM SIGKDD Conference on Knowledge Discovery and Data Mining (KDD '24), August 25--29, 2024, Barcelona, Spain}\acmDOI{10.1145/3637528.3671933}
\acmISBN{979-8-4007-0490-1/24/08}

\makeatletter
\gdef\@copyrightpermission{
 \begin{minipage}{0.3\columnwidth}
  \href{https://creativecommons.org/licenses/by/4.0/}{\includegraphics[width=0.90\textwidth]{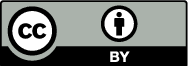}}
 \end{minipage}\hfill
 \begin{minipage}{0.7\columnwidth}
  \href{https://creativecommons.org/licenses/by/4.0/}{This work is licensed under a Creative Commons Attribution International 4.0 License.}
 \end{minipage}
 \vspace{5pt}
}
\makeatother

\usepackage{soul}
\usepackage{url}
\usepackage[utf8]{inputenc}
\usepackage{amsthm}
\usepackage{algorithm}
\usepackage{algorithmic}
\usepackage{multirow}
\usepackage{subcaption}
\usepackage{textcomp}
\usepackage{makecell}
\usepackage{hyperref}

\AtBeginDocument{%
  \providecommand\BibTeX{{%
    \normalfont B\kern-0.5em{\scshape i\kern-0.25em b}\kern-0.8em\TeX}}}

\begin{document}

\title{BTTackler: A Diagnosis-based Framework for Efficient Deep Learning Hyperparameter Optimization}


\author{Zhongyi Pei}
\affiliation{
  \institution{School of Software, BNRist, \\Tsinghua University}
  \city{Beijing}
  \country{China}}
\email{peizhyi@tsinghua.edu.cn}

\author{Zhiyao Cen}
\affiliation{
  \institution{School of Software, BNRist, \\Tsinghua University}
  \city{Beijing}
  \country{China}}
\email{cenzy23@mails.tsinghua.edu.cn}

\author{Yipeng Huang*}
\affiliation{
  \institution{School of Software, BNRist, \\Tsinghua University}
  \city{Beijing}
  \country{China}}
\email{huangyipeng@tsinghua.edu.cn}

\author{Chen Wang}
\affiliation{
  \institution{School of Software, EIRI, \\Tsinghua University}
  \city{Beijing}
  \country{China}}
\email{wang_chen@tsinghua.edu.cn}

\author{Lin Liu}
\affiliation{
  \institution{School of Software, BNRist, \\Tsinghua University}
  \city{Beijing}
  \country{China}}
\email{linliu@tsinghua.edu.cn}

\author{Philip Yu}
\affiliation{
  \institution{School of Software, \\Tsinghua University}
  \city{Beijing}
  \country{China}}
\email{psyu@tsinghua.edu.cn}

\author{Mingsheng Long}
\affiliation{
  \institution{School of Software, BNRist, \\Tsinghua University}
  \city{Beijing}
  \country{China}}
\email{mingsheng@tsinghua.edu.cn}

\author{Jianmin Wang}
\affiliation{
  \institution{School of Software, BNRist, \\Tsinghua University}
  \city{Beijing}
  \country{China}}
\email{jimwang@tsinghua.edu.cn}

\thanks{*Corresponding Author}










\renewcommand{\shortauthors}{Zhongyi Pei et al.}

\begin{abstract}
  Hyperparameter optimization (HPO) is known to be costly in deep learning, especially when leveraging automated approaches.
  Most of the existing automated HPO methods are accuracy-based, i.e., accuracy metrics are used to guide the trials of different hyperparameter configurations amongst a specific search space.
  However, many trials may encounter severe training problems, such as vanishing gradients and insufficient convergence, which can hardly be reflected by accuracy metrics in the early stages of the training and often result in poor performance.
  This leads to an inefficient optimization trajectory because the bad trials occupy considerable computation resources and reduce the probability of finding excellent hyperparameter configurations within a time limitation.
  In this paper, we propose \textbf{Bad Trial Tackler (BTTackler)}, a novel HPO framework that introduces training diagnosis to identify training problems automatically and hence tackles bad trials.
  BTTackler diagnoses each trial by calculating a set of carefully designed quantified indicators and triggers early termination if any training problems are detected.
  Evaluations are performed on representative HPO tasks consisting of three classical deep neural networks (DNN) and four widely used HPO methods.
  To better quantify the effectiveness of an automated HPO method, we propose two new measurements based on accuracy and time consumption.
  Results show the advantage of BTTackler on two-fold:
  (1) it reduces 40.33\% of time consumption to achieve the same accuracy comparable to baseline methods on average and
  (2) it conducts 44.5\% more top-10 trials than baseline methods on average within a given time budget.
  We also released an open-source Python library that allows users to easily apply BTTackler to automated HPO processes with minimal code changes\footnote{https://github.com/thuml/BTTackler}.
\end{abstract}





\maketitle

\section{Introduction}
The significance of hyperparameter optimization (HPO) \cite{bischl_hyperparameter_2023} has been fully acknowledged by deep learning (DL) practitioners as the performance of a deep neural network (DNN) is highly dependent on its hyperparameter configurations.
Since automated approaches emerged to reduce human efforts, HPO has started to build up the core of automated machine learning (AutoML).
With the increasing complexity of DNNs, the training duration and the hyperparameter space continue to expand \cite{yang_hyperparameter_2020}, rendering automated HPO more computationally intensive and time-consuming.

Consequently, research interests have been attracted towards developing more efficient automated HPO methods \cite{tpe,gp,wu_practical_2020,li_hyperband_2017,smac}, while most of them are accuracy-based.
Their typical workflow is to conduct trials of different hyperparameter configurations to train a target model within a predefined search space and a given time limit, aiming to optimize one or several accuracy metrics \cite{agrawal_hyperparameter_2021,hutter_hyperparameter_2019,thornton_auto-weka_2013}.
They generally stop the trials with relatively low accuracy at each epoch or utilize a surrogate model to predict promising hyperparameter configurations to narrow the search.
However, numerous trials may encounter severe training problems, such as exploding gradients, vanishing gradients, abnormal training loss, and insufficient coverage.
With only accuracy-based metrics, these problems are usually difficult to identify in the early stages of the training and result in poor performance or even failure.
Thus, these bad trials can hardly be stopped early and lead to inefficient optimization trajectories due to wasted time and computational resources.

In this work, instead of developing another accuracy-based optimization algorithm, we aim to establish a new perspective on automated HPO.
We propose \textbf{Bad Trial Tackler (BTTackler)}, a novel HPO framework that introduces DNN training diagnosis in automated HPO processes.
BTTackler traces HPO trials and calculates dedicated indicators to detect training problems, thus capable of early terminating problematic training to conserve time and computational resources for more promising trials.
The entire process can be parallelized to minimize the extra overhead.
An open-source Python library has also been developed to facilitate the adoption of BTTackler.

\begin{figure}[htbp]
\centerline
{\includegraphics[width=3.4in]{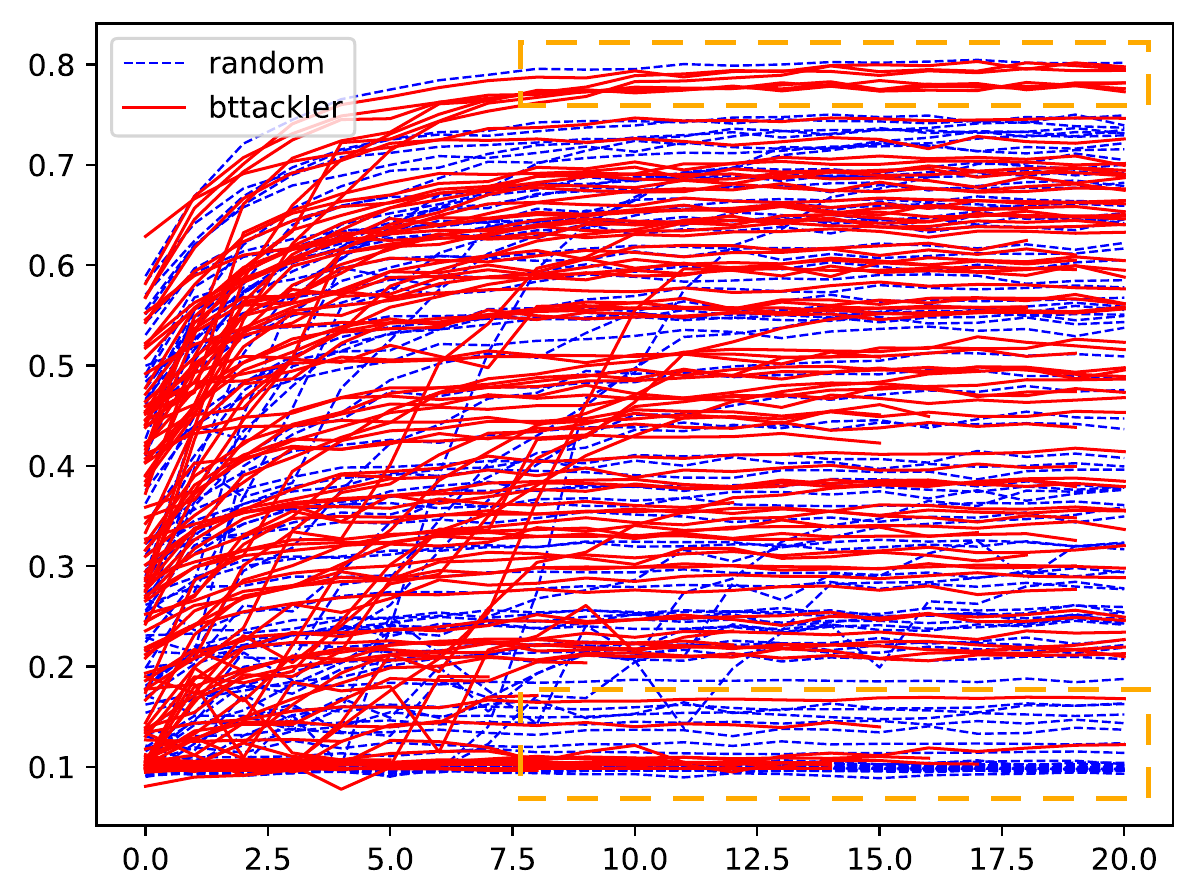}}
\caption{Comparing BTTackler and random search on HPO of a simple CNN on cifar10}
\label{motivation}
\end{figure}

Our motivation can be demonstrated by a preliminary experiment comparing BTTackler with Random Search (RS) \cite{random}, a simple yet effective accuracy-based automated HPO method.
The objective was to optimize the configuration of 13 hyperparameters of a convolutional neural network (CNN) for image classification on $cifar10$ dataset \cite{cifar} within a time limitation of 2 hours using 4 NVIDIA A100 GPUs.
Results are depicted in Figure \ref{motivation}, showing trials by BTTtackler in red lines and those by random search in blue.
From the distribution of blue lines in Figure \ref{motivation}, we can see that all the trials run to the end of the 20th epoch in the HPO of Random Search.
Compared with random search, BTTackler efficiently terminated numerous bad trials (like the red trials in the bottom orange rectangle), finding more hyperparameter configurations with better accuracy (like the red trials in the top orange rectangle).
In the end, BTTackler conducted 134 trials, 30 more than Random Search within the same time budget, resulting in a higher probability of picking up optimal hyperparameter configurations.

In summary, our work makes the following major contributions:
\begin{itemize}
    \item We propose BTTackler, the first HPO framework that introduces training diagnosis into HPO to improve efficiency.
    \item We conduct a literature review of DNN training diagnosis and then design a set of quantified indicators suitable for HPO. 
    \item The evaluation exhibits the significant advantage of BTTackler: (1) for efficiency, it reduces on average 40.33\% of time consumed to achieve the same or similar best accuracy as baseline methods; (2) for performance, it conducts 44.5\% more top-10 trials than baseline methods on average within a given time budget.
\end{itemize}

The rest of the paper is organized as follows: Section \ref{related} compares the related work with BTTackler. Section \ref{approach} describes the main idea and implementation details of BTTackler. Section \ref{experiment} presents the evaluations of BTTackler. Section \ref{conclusion} concludes this paper and discusses possibilities for future work.

\section{Related Work} \label{related}

The related work includes the existing methods of automated HPO and DNN training diagnosis.

\subsection{HPO methods}
HPO is often a non-trivial and delicate task due to the large hyperparameter space and the significant impact of hyperparameters on DNN performance \cite{bischl_hyperparameter_2023}. 
Most existing HPO methods are based on accuracy metrics of the DNN model over a given validation set, such as precision and mean squared error, and fall into three main categories.
The first is Random Search \cite{random}, which selects hyperparameter configurations randomly from the predefined hyperparameter space.
It is straightforward to implement while often achieving competent results in high-dimensional hyperparameter space.
The second category, Bayesian optimization (BO) \cite{shahriari_taking_2016,cho_basic_2020,turner_bayesian_2021}, leverages information learned iteratively from the past trials of the HPO task to prioritize the hyperparameter searching. 
The third category, multi-fidelity methods \cite{li_hyperband_2017,falkner2018bohb,eggensperger_hpobench_2022}, focuses on efficiently allocating resources to promising hyperparameter configurations and terminating trials with poor performance to speed up the HPO process.

In addition to the above HPO methods, early termination rules (ETRs) \cite{medianstop,curvefitting} have also been proposed to improve HPO efficiency.
These rules make certain assumptions of DNN learning curves based on accuracy metrics and terminate poor-performance trials.
Early termination is widely used, but the reliability issue of such methods remains open in both academic research and practice.
Due to the instability of accuracy metrics during training, good configurations may also perform badly at the early stages of the training, risking elimination by ETRs.
Instead of building upon accuracy assumptions, BTTackler leverages early termination through training diagnosis to improve the efficiency of HPO.

\subsection{Training Diagnosis}\label{background-training}
DNN training diagnosis is closely related to DNN model testing.
DeepXplore \cite{pei_deepxplore_2017} is a white-box framework for systematical testing of real-world deep learning systems, introducing neuron coverage as an indicator of model health.
DiffChaser \cite{xie_diffchaser_2019} is an automated genetic algorithm-based testing technique to discover disagreement of multiple DNN version variants using prediction uncertainty indicators.
CKA \cite{kornblith2019similarity_CKA_Hinton} identifies correspondences between representations in DNNs, which may indicate some potential training issues.
MEGAN \cite{xiong2023model_megan} is proposed to predict the generalization performance of DNNs by utilizing the mean empirical gradient norms of the last layer.
These methods generally work in the post-training stages and consume lots of computational resources, which can hardly be used in HPO as they increase massively the performance overhead.

Other works were proposed to diagnose the training of DNNs and localize potential faults.
UMLUAT \cite{schoop_umlaut_2021} is an interactive tool enabling users to check the model structure and metrics during DNN training, providing messages of potential bugs and corresponding repair methods.
DeepLocalize \cite{wardat_deeplocalize_2021} identified the faulty layer by detecting the numerical errors of the neurons.
Amazon SageMaker Debugger \cite{rauschmayr_amazon_2021} provided a set of built-in heuristics to debug faults in DNN training.
More practical training diagnosis methods have also been proposed in recent years.
AUTOTRAINER \cite{zhang_autotrainer_2021} summarized five common training problems and proposed a systematic diagnosis method suitable for various DNN architectures.
DeepDiagnose \cite{wardat_deepdiagnosis_2022} defined eight training issues collected from GitHub and Stack Overflow.
DeepFD \cite{cao_deepfd_2022} is a learning-based fault diagnosis and localization framework that maps the fault diagnosis task to a multi-classification problem using fault seeding.

While the existing methods offer various solutions for detecting training problems, attempts to apply them in automated HPO are complex and challenging.
The motivation difference between training diagnosis and HPO is significant.
The former generally intends to correct issues in the code or data of machine learning tasks, while these issues should be resolved before HPO.
The bounds and thresholds from the previous research are also unsuitable for HPO. In those methods, training diagnosis is often followed by repairing. 
They are quite tolerant of misdiagnosis because it is almost harmless to repair when training is not problematic. 
However, indulged misdiagnoses could lead to missing many good hyperparameters when using training diagnosis to terminate trials in automated HPO.

In addition, some of the methods that provide descriptions and suggestions for training problems may not be applicable since they require human intervention and thus block automated HPO processes.
Moreover, the computational cost highly impacts the efficiency of automated HPO.
As a result, training diagnosis methods that generate excessive overhead should be avoided in BTTackler.

\section{Diagnosis-based HPO Framework} \label{approach}
BTTackler must carefully extract knowledge from the existing training diagnosis methods due to the motivation difference presented above.
The primary challenge is determining the actual training problems pertinent to automated HPO.
We first introduce \textit{Quality Indicator} to define the selected quantified expertise to elucidate their functionality in HPO.
\begin{definition}[\textit{Quality Indicator}]
A quality indicator is a quantified method for identifying specific training problems that may lead to failure or poor performance in DNN training.
\end{definition}
Some well-known examples of training problems include exploding gradients, vanishing gradients, and abnormal training loss values. 
We further define \textit{Diagnosis-based HPO} to distinguish our proposed framework from the existing accuracy-based HPO. 
\begin{definition}[\textit{Diagnosis-based HPO}]
Diagnosis-based HPO introduces training diagnosis into existing HPO methods. It utilizes quality indicators to detect training problems and terminate bad trials to achieve higher accuracy and efficiency.
\end{definition}

\begin{figure*}[htbp]
  \centering
  \includegraphics[width=6.3 in]{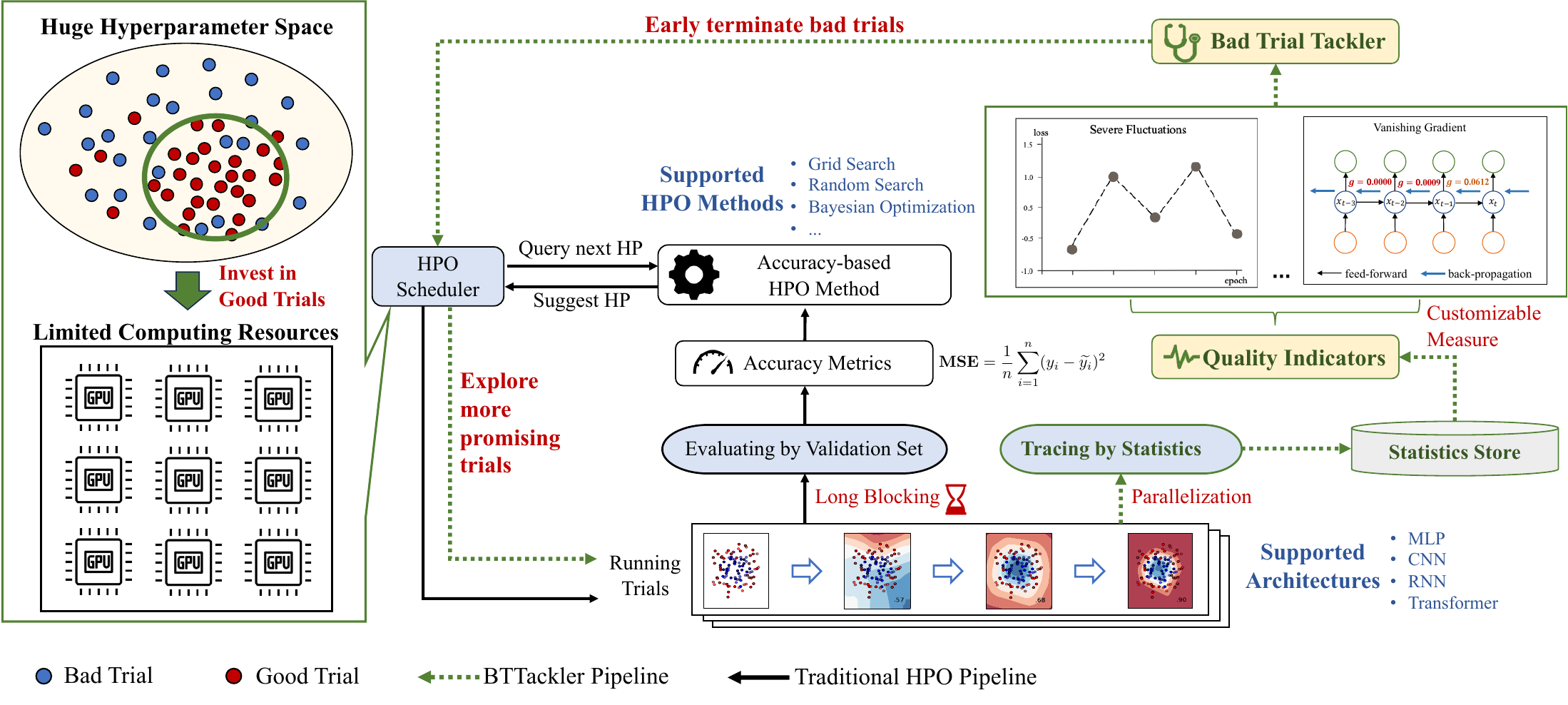}
  \caption{The framework of BTTackler.}
  \label{arch}
\end{figure*}

In developing DNN models, HPO is an integral and highly costly step.
In HPO, hundreds or even thousands of hyperparameter configurations must be tried to find an optimal one, which requires massive computing resource consumption.
Many experts are used to manually terminating hopeless experiments as soon as possible to save computing resources for more promising trials.
However, such implicit knowledge remains in the experts' heads, and due to manual operations' low efficiency, they generate little benefit to the whole HPO task.
During the past few years, training diagnosis has become a rising research focus in the field of software engineering\cite{wardat_deepdiagnosis_2022,zhang_autotrainer_2021,ma_mode_2018,eniser_deepfault_2019}. 
Previous works identified some symptoms that may lead to training failure or poor performance of DNNs.
These works bring a possible alternative to manual termination in HPO.
In this paper, we propose an HPO framework for DNNs, BTTackler, that automatically tackles bad trials using the training problems identified by previous research findings.
The key idea and framework of BTTackler is illustrated in Figure \ref{arch}.

\subsection{The BTTackler Pipeline}

We first presented the traditional HPO pipeline in black arrowed lines in Figure \ref{arch}.
The HPO scheduler can be seen as the executor of HPO trials. It queries hyperparameter configurations from the HPO methods and invokes new processes to run trials with the suggested configurations.
Limited by the amount of computing resources, the concurrency of HPO is usually very low. So, only when previous trials are finished can the subsequent trials be executed by the HPO scheduler.
Because of the vast hyperparameter space, an HPO task may contain hundreds or thousands of trials.
With a limited time budget, there are two ways to achieve better effectiveness of HPO. 
One is to pick up potentially better hyperparameter configurations by modeling the relationship between hyperparameters and the performance of DNN models, like BO methods.
The other is to terminate relatively poor-performing trials by comparing them simultaneously, like fidelity-based HPO methods.
However, neither of the two approaches considers expertise in identifying training problems, which is usually used during manual HPO by experts in practice.

Based on the traditional HPO pipeline, BTTackler brings a novel branch that automatedly terminates bad trials with some expertise applicable to HPO.
The primary challenge to BTTackler is quantifying expertise to support decision-making in the HPO pipeline.
We name the quantified expertise as quality indicators, indicating whether there are problems with training.
With the rise of training diagnosis research, some methods were proposed to detect training problems.
We conduct a literature review of training diagnosis and construct quality indicators for HPO, as presented in Appendix \ref{literature}.
In the BTTackler pipeline, there are two main processes: one is the process $Tracer$ that traces trials, and the other one is the process $Checker$ that utilizes quality indicators to detect training problems.
The whole procedure can be viewed in two stages.
In the first stage, $Tracer$ monitors the trial by analyzing the statistics of variables that reflect training problems, such as training losses, gradients, weights, etc.
All the statistics are stored in files by $Tracer$ and then conducted to $Checker$.
In the second stage, $Checker$ calculates the quality indicators and signals the HPO scheduler to stop trials when training problems are indicated.
Efficiency is critical in HPO, so the BTTackler pipeline is parallelized as much as possible, which is presented in detail in Section \ref{implementation}.

\subsection{Quality Indicators}
\begin{figure*}[htbp]
  \centering
  \includegraphics[width=6.3 in]{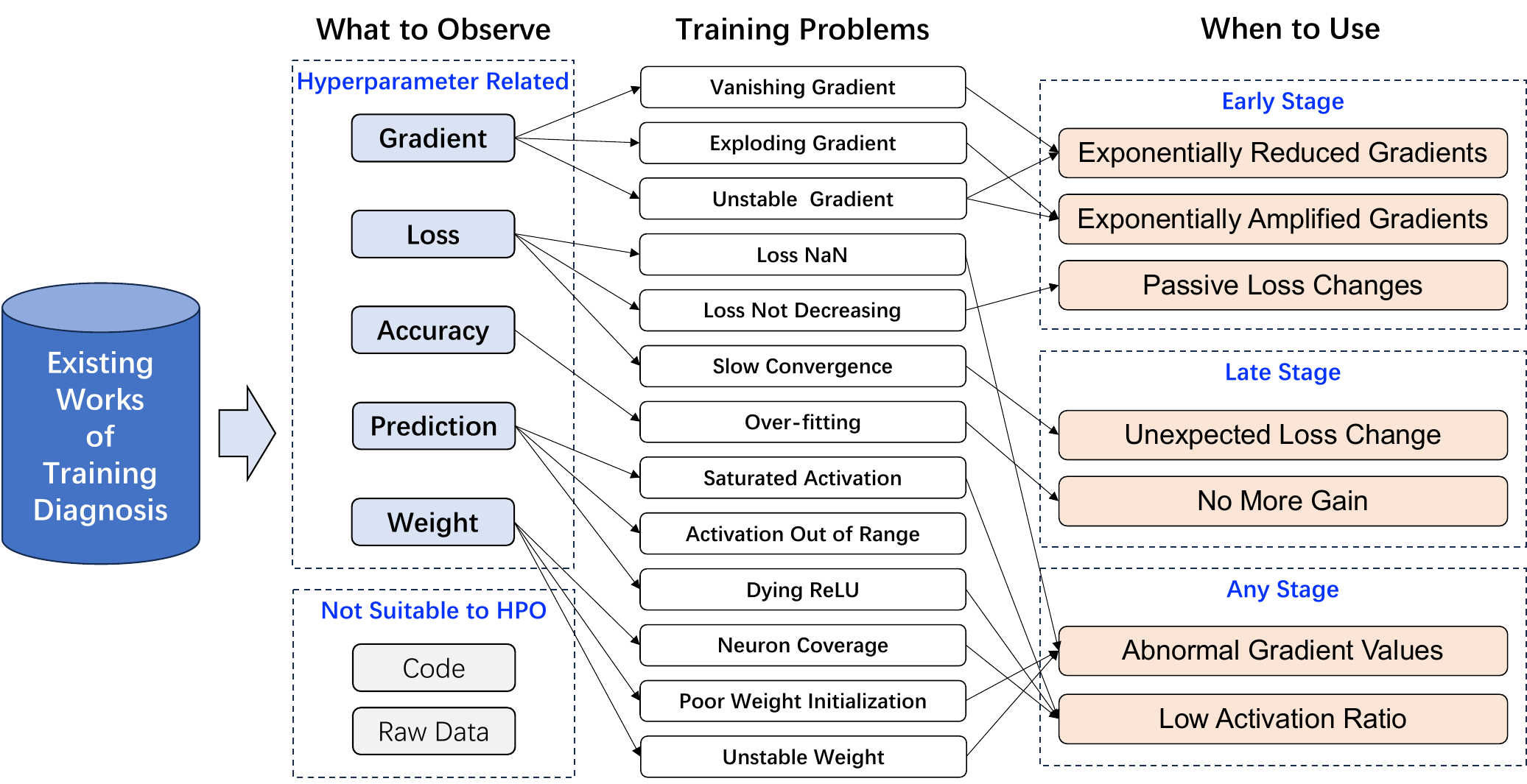}
  \caption{The Design of Quality Indicators.}
  \label{indicators_fig}
\end{figure*}

As a summary, we describe the logic of developing quality indicators in Figure \ref{indicators_fig}.
At the very first step, we finished a brief literature review of training diagnosis of DNNs to find available and robust evidence to detect training problems, as Appendix \ref{literature} shows. 
After learning from the work of training diagnosis, we propose seven quality indicators for the BTTackler pipeline.

\subsubsection{What concretely are quality indicators? } \label{what}
Like the training diagnosis methods, quality indicators keep observing some variables during training and alert problems when specific symptoms emerge.
However, not all of those can be used in quality indicators, such as code and raw data, since HPO is the process of tuning hyperparameters instead of debugging codes or validating data.
Each observed variable must be quantified to support programming to alert training problems, which is implemented primarily by manually defined rules.
We describe the proposed seven quality indicators as follows:

\textbf{\textit{Abnormal Gradient Values (AGV).}}
We use AGV to denote the quality indicator for detecting abnormal gradients during back-propagation. 
Abnormal gradients can severely impact the training trend and result in poor performance or undesirable outcomes. 
The definition of AGV involves gradient values that are not-a-number (\texttt{nan}), infinite (\texttt{inf}), or violate common sense. 
Specifically, two rules are used for AGV. 
First, any value of gradients must not be \texttt{nan} or \texttt{inf}.
The second is that the absolute values of gradients for each layer of DNNs must be smaller than an empirical safe upper bound. 
When any of the two rules are violated, AGV becomes positive.

\textbf{\textit{Exponentially Amplified Gradients (EAG).}}
We use EAG to denote an early warning for potential exploding gradients during training. 
Exploded gradients are usually generated from repeated amplifications of gradients.
So, we evaluate the overall amplifications of gradients between adjacent layers of DNNs as an approximate of EAG.
Specifically, the overall median of the amplifications of gradients between adjacent layers of DNNs must be smaller than an empirical safe upper bound; otherwise, EAG becomes positive.

\textbf{\textit{Exponentially Reduced Gradients (ERG).}}
We use ERG to indicate early signs of vanishing gradients. 
ERG employs a similar computation procedure to EAG. 
It explicitly identifies exponential drops in gradient magnitudes between adjacent layers of DNNs.
The overall median of the amplifications of gradients between adjacent layers of DNNs must be more significant than an empirical safe lower bound; otherwise, ERG becomes positive.

\textbf{\textit{Passive Loss Changes (PLC).}} 
We use PLC to denote the weak effect problem in the early stages of training. 
The changes in training loss should be noticeable when the hyperparameters about optimization are well set. 
Conversely, the trial will probably result in a bad outcome when training loss does not exhibit the expected decreasing trend. 
We collect the training loss values for each epoch to monitor this behavior. 
Then, the absolute values of training loss change between adjacent epochs are calculated, denoted as $[G_{i}]$, where the gap is from epoch $i$ to epoch $i+1$. 
At last, we calculate the ratio of the mean of $[G_{i}]$ to the initial training loss as the criteria. When it is not noticeable, such as smaller than $0.1\%$, PLC is positive.

\textbf{\textit{Low Activation Ratio (LAR).}} 
We use LAR to denote the problem that considerable neurons are inactivated during training.
A neuron is activated if its output exceeds a threshold (e.g., 0) and varies in a reasonable range.
If too few neurons are activated, i.e., the activation ratio is low, it potentially results in serious gradient problems during the back-propagation of training.
We calculate the ratio of zero neurons in each layer to detect such issues and compare it with an empirical upper threshold from prior research. 
LAR is positive if the ratio is more significant than the empirical threshold.

\textbf{\textit{Unexpected Loss Changes (ULC).}} 
We use ULC to denote unexpected trends of training loss in the later stages of training.
The training loss should follow a converging trend in the later stages rather than intense fluctuations or increases. 
Such unexpected trends indicate that the training is not stable.
We employ linear regression to quantify the fluctuations within an adaptive window whose size varies with the maximum training epoch. 
The fluctuations must be smaller than a tolerance threshold derived from previous research; otherwise, ULC is positive. 

\textbf{\textit{No More Gain (NMG).}} 
We use NMG to denote a benign issue that further training should be unnecessary. 
For a good trial, training loss and accuracies will finally converge in the later stages of training, indicating a very low possibility of further improvement of training. 
With NMG, we can identify such situations and recommend early termination to save computing resources.
We collect the minimum of the latest training losses within an adaptive window whose size varies with the maximum training epoch. 
If the minimum within the window exceeds the overall minimum during training, NMG is positive.

Introducing some empirical parameters in the definitions of quality indicators is inevitable because the expertise in detecting training problems is not static or immutable.
Theoretically, we can tune those empirical parameters to make the quality indicators suitable to a specific domain considering neural network architectures and tasks.
In this paper, we intend to show BTTackler's generalizability.
So, notably, the quality indicators are designed to be conservative, reflecting on the choices of relatively easy threshold values and loose rules.
This means that the quality indicators can only detect severe training problems, which helps reduce the risk of mistaking good hyperparameters for bad ones.

\subsubsection{How to use quality indicators? } 
A quality indicator must be used at the right time to support the BTTackler pipeline.
For example, loss is expected to change intensively when training begins, but it is probably problematic when training is nearing its end. 
Therefore, applying quality indicators should be within a suitable scope of training. 
To simplify this issue, we split training into two stages, i.e., early stage and late stage.
ERG, EAG, and PLC are used at the early training stage because the related training problems are more remarkable initially.
Meanwhile, ULC and NMG are used at the late stage of training to check that training is already enough.
AGV and LAR can be used at any stage since they happen randomly.

To exhibit BTTackler's generalizability, we use all the above quality indicators in parallel for several representative cases of DNN training.
During each trial in HPO, when any quality indicator at its working stages is positive, BTTackler triggers early termination.
Note that the alerted trials will be terminated but not eliminated for the quality indicators indicating training enough problems because such hyperparameter configurations may also be candidates for the best ones.

\subsection{Implementation} \label{implementation}

The pipeline of BTTackler comprises recording training variables, parallel calculating quality indicators, and early terminations.


\textbf{Recording Training Variables.} 
Considering the characteristics of various training environments, we adopt two approaches to trace the variables of training. 
The first approach periodically records runtime information, such as model weights, training loss, and validation accuracy, by inserting predefined probes into the training procedure's codes. 
The second approach utilizes the widely used callback functions of deep learning libraries. 
The callback functions can record variables that are not easily observed outside the deep learning libraries.
Because of the vast amount of data, storing every neuron's weights and gradients for each training iteration would be undesirable. To address this, we adopt an intuitive approach from existing works \cite{wardat_deeplocalize_2021,cao_deepfd_2022,zhang_autotrainer_2021}, storing only the fundamental statistical values of the variables. 
To ensure the extensibility to support various quality indicators, we store ten representative statistics for each variable, including \textit{average}, \textit{variance}, \textit{median}, \textit{minimum}, \textit{maximum}, \textit{upper quartiles}, \textit{lower quartiles}, \textit{skewness}, \textit{kurtosis}, and \textit{ratio of zero}.
Specifically, DNNs' gradients and weights are stored layer-wise and epoch-wise, while the loss value and accuracy are stored epoch-wise. 

\textbf{Parallelization and Overhead.} 
The quality indicators' runtime overhead must be considered, which is one of the criteria for selecting useful references for implementing quality indicators in Appendix \ref{literature}.
We also utilize multi-threading libraries to implement a two-level parallelization to avoid blocking the model training process. 
On the one hand, the calculation process is isolated from the primary process to avoid blocking training or evaluation.
On the other hand, we make each quality indicator run on its thread to improve efficiency and reduce the impact of unexpected errors from any quality indicator.

\textbf{Early Terminations}. 
BTTackler conducts early termination when quality indicators report training problems. 
On one side, a reporter collects the results from the adopted quality indicators for a given trial. It sends an early termination query once any of the results are positive.
On the other side, the HPO scheduler of BTTackler keeps listening to queries about early termination and kills the process of the trials being required to terminate.
Notably, each trial's final performance will be used to update the surrogate model in the Bayesian methods of HPO. 
In the popular open-source library NNI \cite{nni}, when early termination is triggered (by accuracy-based methods) in Bayesian HPO methods, the last evaluation result is used as the final performance. 
Since early termination denotes a pessimistic estimation, the last evaluation could represent the model's performance with a small risk of information losses. 
BTTackler followed this way. 

\textbf{Simulator for Calibration.} \label{simulation}
To efficiently calibrate quality indicators in BTTackler, we propose a simulator-based method to evaluate the quality indicators without really conducting HPO trials.
As a preparation, a representative real HPO task is executed, and all the information needed by BTTackler is recorded.
Then, developers could evaluate their customized quality indicators by replaying the recorded HPO task.
During the replay, the simulator sequentially picks the records of trials generated by the HPO method and calls BTTackler to diagnose the training. 
In this way, the definition of quality indicators, including the empirical parameters, can be modified according to the effectiveness and generalizability of the simulation, as the goal we presented in Section \ref{what}.

\subsection{Evaluation Method}

We can evaluate the effectiveness of BTTackler in two ways.
The first is to compare the accuracy metrics of an HPO method and its BTTackler-assistant version at a fixed time budget.
Considering the randomness in HPO, the measurement of the accuracy gap is unreliable.
To achieve a stable and fair comparison, we define a new measurement for HPO methods using the ratio of top-accuracy trials.
\begin{definition}[\textit{Top10 Hit Ratio}] \label{top10hr}
To compare an HPO method $i$ and its baseline method $j$, we first collect the overall top 10 trials on accuracy metrics from the two methods.
Top10 Hit Ratio (Top10HR) is the ratio of the trials from the HPO method $i$, which is denoted by $\text{Top}10\text{HR}_{ij}=\frac{K_i}{10} * 100\%$, where $K_i$ is the number of the trials from the HPO method $i$ that rank in the overall top 10 trials.
\end{definition}

We use top $10$ in Top10HR because HPO targets the best trials, and top $10$ is an appropriate range considering randomness.
The second way to evaluate the effectiveness of BTTackler is to measure the time cost saving at the point where the best accuracy of the baseline method is achieved.
\begin{definition}[\textit{Time-Saving for Baseline Accuracy}]
To compare an HPO method $i$ and its baseline method $j$, we first record the best accuracy $A_j$ of the baseline method $j$ given a fixed time budget $T_j$. 
Time-saving for Baseline Accuracy (TSBA) is the relative reduction in time cost at the point when the HPO method $i$ reaches $A_j$, which is denoted by $\text{TSBA}_{ij} = \frac{T_j - T_i}{T_j} * 100\% $.
\end{definition}

\section{Experiment} \label{experiment}

In this section, we study three research questions to evaluate BTTackler:
\begin{itemize}
    \item [\textbf{RQ1}] What is the effectiveness of BTTackler on the accuracy?
    \item [\textbf{RQ2}] How does BTTackler improve the efficiency of HPO?
    \item [\textbf{RQ3}] What are the individual impacts of quality indicators?
\end{itemize}

\subsection{Setup}
We implemented BTTackler based on the popular open-source HPO framework \textit{NNI} \cite{nni}.
All experiments are conducted on three homogeneous servers of \textit{Intel(R) Xeon} $14$-core processors, $384$GB of RAM, and $8$ \textit{NVIDIA TITAN X (Pascal)} GPUs, and \textit{Ubuntu 18.04}.
The concurrency of HPO experiments was uniformly set at $8$.
We performed our evaluation on three distinct and representative commonly used DNN architectures. The HPO tasks are summarized as follows:

\begin{itemize}
    \item \textbf{\textit{Cifar10CNN:}} The \textit{Cifar10}\cite{cifar} dataset consists of $60000$ $32\times32$ color images in $10$ classes. The CNN architecture contains $4$ Conv layers and $3$ MLP layers. This classification task has 3 continuous, 5 discrete, and 5 categorical hyperparameters.
    \item \textbf{\textit{Cifar10LSTM:}} \textit{LSTM}\cite{LSTM} is short for long short-term memory recurrent neural network. This classification task has 3 continuous, 1 discrete, and 4 categorical hyperparameters.
    \item \textbf{\textit{Ex96Trans:}} \textit{Exchange}\cite{exchange} is a time-series dataset that records the daily exchange rates of 8 countries from $1990$ to $2016$. 
    \textit{Transformer} \cite{vaswani2017attention} is one of the most successful DNN models in recent years, but its difficulty in training presents challenges to HPO. This forecasting task has 1 continuous, 4 discrete, and 3 categorical hyperparameters.
\end{itemize}

We list the three tasks' hyperparameter space in Table \ref{hp},
where \textit{HP-$I$} denotes continuous hyperparameters, \textit{HP-$II$} denotes discrete hyperparameters, and \textit{HP-$III$} denotes categorical hyperparameters.
In order to make the HPO sufficiently challenging, we set large search spaces of hyperparameter configurations for all the tasks.
Details can be seen in the open-source project.

\begin{table}[htbp]
\caption{Number of different hyperparameters.}
\begin{center}
\begin{tabular}{lcccc}
\toprule
\textbf{Setup} & \textbf{HP-$I$}&\textbf{HP-$II$}& \textbf{HP-$III$} & \textbf{Task Type}\\
\midrule
Cifar10CNN              & 3 & 5 & 5 & classification \\ \midrule
Cifar10LSTM             & 3 & 1 & 4 & classification \\ \midrule
Ex96Trans   & 1 & 4 & 3 & forecasting \\
\bottomrule
\end{tabular}
\label{hp}
\end{center}
\end{table}

We evaluated four representative HPO methods as baselines that are used in the well-known open-source HPO framework\cite{nni}, including \textit{Random Search (RS)} \cite{random}, \textit{Gaussian Process (GP)} \cite{gp}, \textit{Tree structure Parzen Estimator (TPE)} \cite{tpe} and \textit{ Sequential Model-Based Optimization for General Algorithm Configuration (SMAC)} \cite{smac}.
Due to its simplicity, we selected \textit{RS} as the baseline. 
\textit{GP} was chosen for its surrogate modeling capabilities.
\textit{TPE} was included for its advanced Bayesian optimization strategies, particularly effective in complex HPO scenarios. 
Lastly, \textit{SMAC} was chosen as a state-of-the-art method for categorical hyperparameters, setting a high standard for efficiency and accuracy in HPO.
We followed the settings of the HPO methods in \textit{Deep-BO}\cite{deep-bo}.

Our comparative experiments also consider two early termination rules (ETRs) as baselines, namely Learning Curve Extrapolation (LCE) and Median Stop Rule (MSR).
\textit{LCE} \cite{curvefitting} stops a trial if its forecast accuracy at the target step is lower than the best performance in history, used in the classification task in Cifar10CNN and Cifar10LSTM.
\textit{MSR} \cite{medianstop} stops a trial if the trial’s performance is worse than the median value of all completed trials’ performance at the current step, used in the forecasting task in Ex96Trans.
Both LCE and MSR are popular and well-supported methods in improving the efficiency of hyperparameter optimization without sacrificing model performance\cite{curvefitting,medianstop}.
The augments of LCE and MSR follow the default settings in NNI\cite{nni}. 


\subsection{Effectiveness (RQ1)}

\begin{table*}[htbp]
\caption{The effectiveness of BTTackler-enhanced HPO methods on three tasks.}
\begin{center}
\begin{tabular}{l|ccc|ccc|ccc}

\toprule

\textbf{Task}&\multicolumn{3}{|c|}{\textbf{Cifar10CNN}} 
& \multicolumn{3}{|c|}{\textbf{Cifar10LSTM}} 
& \multicolumn{3}{|c}{\textbf{Ex96Trans}} 
\\
\midrule
\textbf{Evaluation Metric}&\multicolumn{3}{|c|}{\textbf{Classification Accuracy}} 
& \multicolumn{3}{|c|}{\textbf{Classification Accuracy}} 
& \multicolumn{3}{|c}{\textbf{MSE}} 
\\
\midrule

\textbf{Performance} & \textbf{Top1}& \textbf{Top10}& \textbf{Top10HR}& \textbf{Top1}& \textbf{Top 10}&\textbf{Top10HR}& \textbf{Top1}& \textbf{Top10} & \textbf{Top10HR}\\
\midrule
Random & 0.7124 & 0.6863 & - & \textbf{0.5488} & 0.5077 & - & \textbf{0.1857} & 0.1899 & - \\ 
Random-ETR & 0.6635 & 0.6415 & 51\% & 0.5407 & \textbf{0.5210} & \textbf{56\%} & 0.1868 & \textbf{0.1894} & 62\% \\
\textbf{Random-BTTackler} & \textbf{0.8200} & \textbf{0.7980} & \underline{\textbf{97\%}} & 0.5337 & 0.5040 & 35\% & \textbf{0.1857} & \textbf{0.1894} & \textbf{64\%} \\ \midrule
GP & 0.7605 & 0.7294 & - & \textbf{0.5522} & \textbf{0.5268} & - & \textbf{0.1851} & 0.1899 & - \\ 
GP-ETR & 0.6771 & 0.6734 & 32\% & 0.5163 & 0.4890 & 16\% & 0.1853 & 0.1894 & 74\% \\
\textbf{GP-BTTackler} & \textbf{0.8040} & \textbf{0.7850} & \textbf{86\%} & 0.5357 & 0.5226 & \textbf{24\%} & \textbf{0.1851} & \textbf{0.1841} & \underline{\textbf{76\%}} \\ \midrule
TPE & 0.6818 & 0.6544 & - & 0.4989 & 0.4720 & - & 0.1854 & 0.1890 & - \\ 
TPE-ETR & 0.7393 & 0.7047 & 60\% & 0.4921 & 0.4328 & 65\% & \textbf{0.1825} & 0.1890 & 57\% \\
\textbf{TPE-BTTackler} & \textbf{0.8108} & \textbf{0.7910} & \textbf{95\%} & \textbf{0.5222} & \textbf{0.5039} & \textbf{86\%} & 0.1837 & \textbf{0.1855} & \textbf{77\%} \\ \midrule
SMAC & 0.8390 & 0.8347 & - & 0.5580 & 0.5396 & - & 0.1826 & 0.1885 & - \\ 
SMAC-ETR & 0.6804 & 0.6715 & 17\% & 0.5500 & 0.5333 & 63\% & 0.1805 & 0.1852 & 72\% \\
\textbf{SMAC-BTTackler} & \underline{\textbf{0.8399}} & \underline{\textbf{0.8367}} & \textbf{57\%} & \underline{\textbf{0.5654}} & \underline{\textbf{0.5541}} & \underline{\textbf{94\%}} & \underline{\textbf{0.1802}} & \underline{\textbf{0.1838}} & \underline{\textbf{76\%}} \\

\bottomrule
\end{tabular}
\label{effective}
\end{center}
\end{table*}

\begin{figure*}[htbp]
\centering
    \begin{subfigure}[b]{0.33\textwidth}
         \centering
         \includegraphics[width=\textwidth]{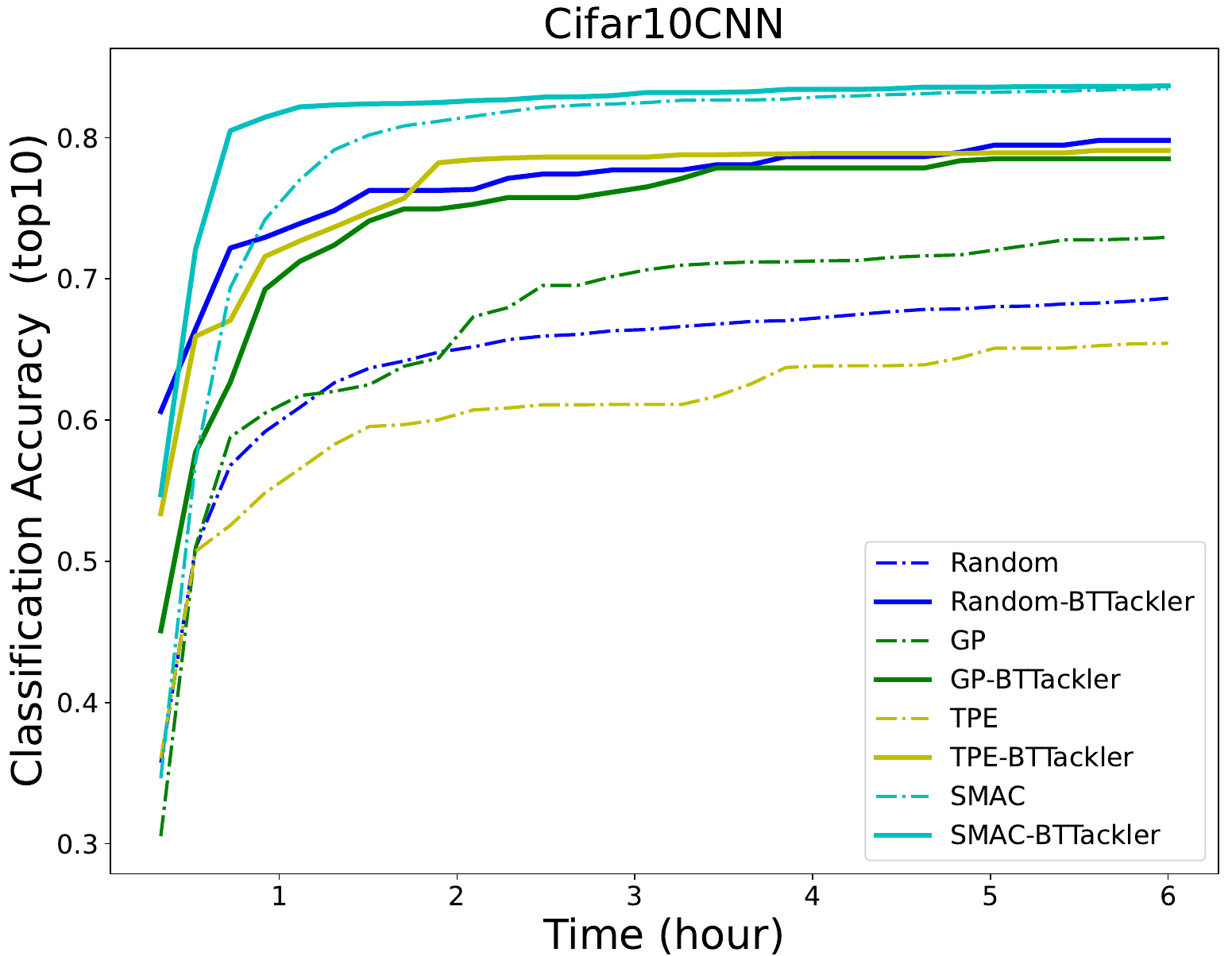}
         \caption{Cifar10CNN}
         \label{1}
    \end{subfigure}
    \begin{subfigure}[b]{0.33\textwidth}
         \centering
         \includegraphics[width=\textwidth]{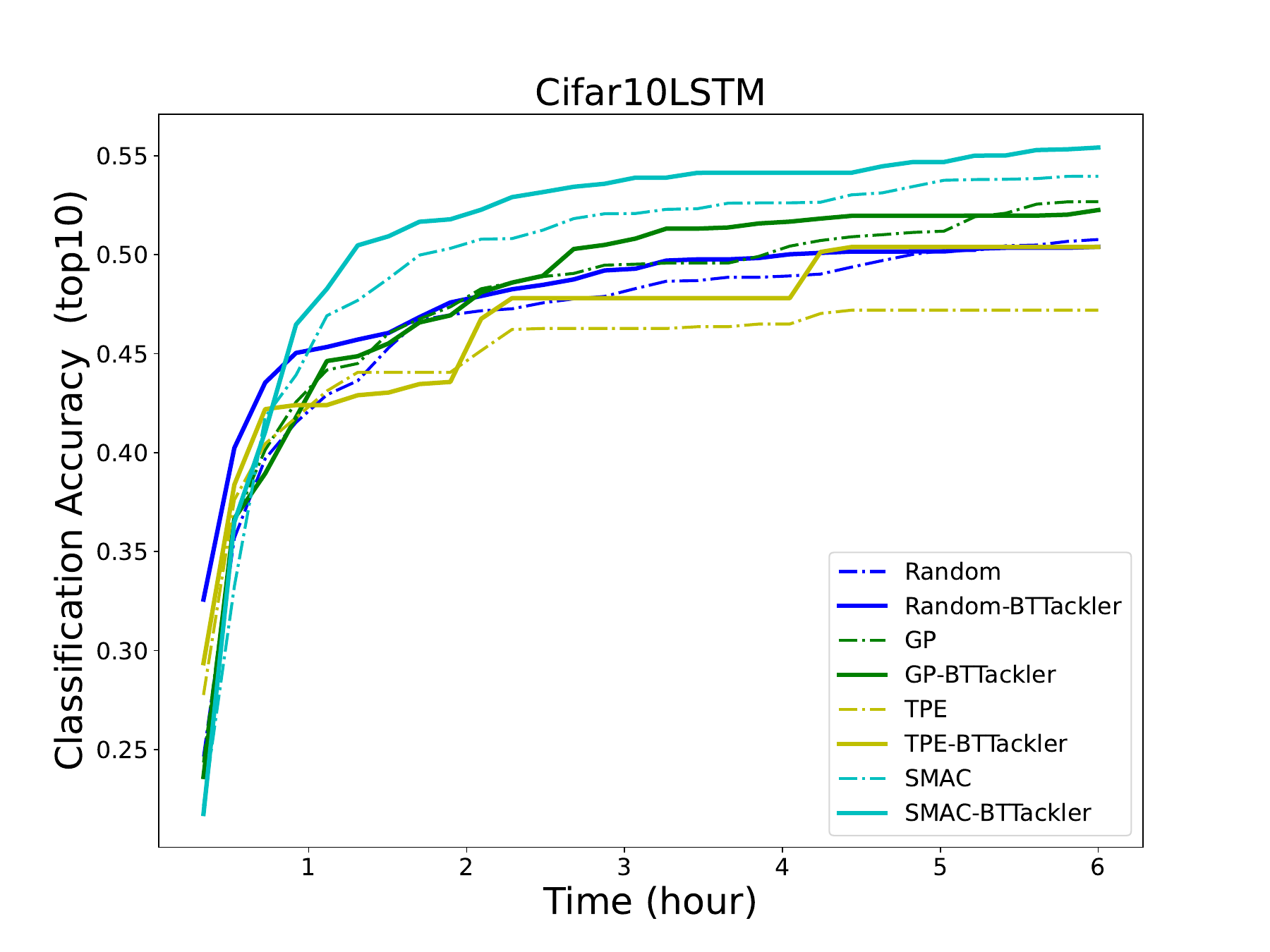}
         \caption{Cifar10LSTM}
         \label{2}
    \end{subfigure}
    \begin{subfigure}[b]{0.33\textwidth}
         \centering
         \includegraphics[width=\textwidth]{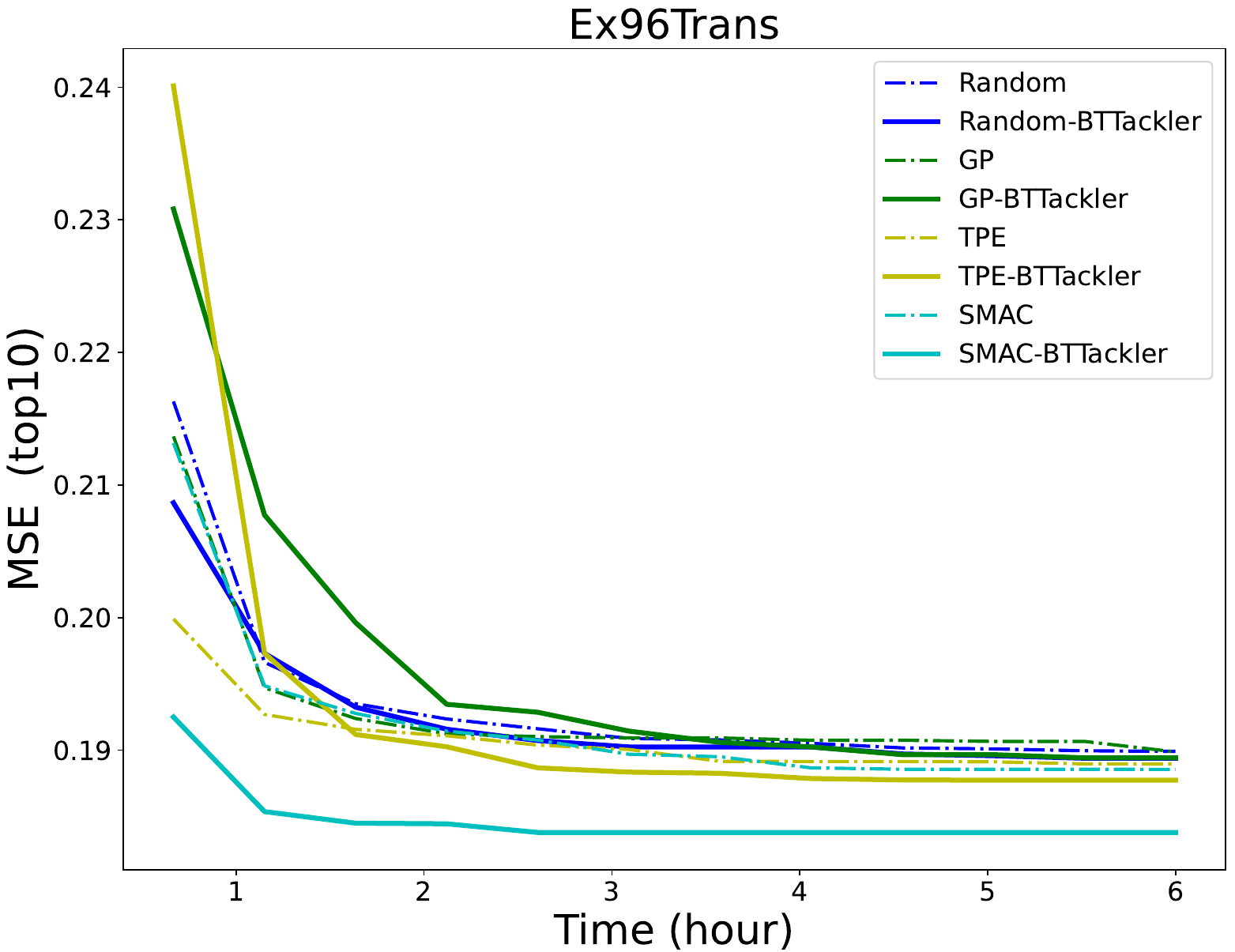}
         \caption{Ex96Trans}
         \label{3}
    \end{subfigure}
\caption{The effectiveness of BTTackler-enhanced HPO methods over time.}
\label{fig-effective} 
\end{figure*}

To evaluate the effectiveness of BTTackler, we ran all four baselines of HPO methods over three tasks and compared them with the BTTackler-enhanced versions, as shown in Table \ref{effective}.
For each baseline $x$, we use \textit{$x$-BTTackler} to denote the BTTackler-enhanced version.
The time budget for all the experiments was $6$ hours.
To mitigate the randomness of training, we repeated all experiments 3 times for each case and used the average as the results.
In view of accuracy, we compare the average performances of the top $1$ and top $10$ trials to show the accuracy gain from BTTackler, considering randomness and fairness.
However, the accuracy gap does not directly reflect the advantage of HPO methods because HPO methods search hyperparameters instead of optimizing models. 
A reasonable comparison is Top10HR from Definition \ref{top10hr}.
Top10HR is measured on $x$-BTTackler and $x$-ETR based on $x$ to compare the effectiveness of ETRs and BTTackler in a way closer to practical usage.
If Top10HR is bigger than $50\%$, \textit{$x$-}method dominates the baseline $x$, and a bigger Top10HR means a greater advantage.
We use bold to denote the better performance between the baseline $x$ and the BTTackler-enhanced version \textit{$x$-BTTackler}. Besides, the best result for each column is underlined.

The \textit{$x$-BTTackler} methods surpassed the baseline $x$ in most cases on average and significantly improved each task's best performance.
BTTackler achieved accuracy gains in all comparisons of \textit{Cifar10CNN} and \textit{Ex96Trans}. 
Whereas the two baselines, \textit{Random} and \textit{GP} in \textit{Cifar10LSTM}, were not beaten by BTTackler.
As the training of \textit{LSTM} is usually unstable, there is high occasionality in results.
We further present the experiment results over time in Figure \ref{fig-effective}.
The \textit{$x$-BTTackler} methods generally took advantage after 1h or 2h and kept the advantage for the most time.
Even for the two bad cases in \textit{Cifar10LSTM}, \textit{Random-BTTackler} and \textit{GP-BTTackler} performed better from 2h to 5h.
Though \textit{SMAC} exhibited high stability as it outperformed all the HPO methods, obvious performance gains were still achieved by the BTTackler-enhanced version.
The average values of Top10HR are $72.25\%$ for BTTackler and $52.08\%$ for ETRs, demonstrating that BTTackler outperforms ETRs significantly.


\subsection{Efficiency (RQ2)}

\begin{table*}[htbp]
\caption{The efficiency of BTTackler-enhanced HPO methods on three tasks. The numbers in the table represent the total number of hyperparameter configurations tried in each experiment.}
\begin{center}
\begin{tabular}{m{2.6cm} |m{1.15cm} m{1.15cm} m{1.15cm}| m{1.15cm} m{1.15cm} m{1.15cm}| m{1.15cm} m{1.15cm} m{1.15cm}  }
\toprule
\textbf{Task}&\multicolumn{3}{|c|}{\textbf{Cifar10CNN}} 
& \multicolumn{3}{|c|}{\textbf{Cifar10LSTM}} 
& \multicolumn{3}{|c}{\textbf{Ex96Trans}} 
\\
\midrule
\textbf{Time Budget} & \textbf{1h}&\textbf{3h}& \textbf{6h}& \textbf{1h}&\textbf{3h}& \textbf{6h}& \textbf{1h}&\textbf{3h}& \textbf{6h} \\
\midrule
Random  & 63 & 186 & 367 & 62 & 180 & 356 & 61 & 163 & 307 \\ \midrule
Random-ETR  (\textbf{increase ratio})  & 105 (\textbf{66.7\%$\uparrow$}) & 323 (\textbf{73.7\%$\uparrow$}) & 662 (\textbf{80.4\%$\uparrow$}) & 100 (\textbf{61.3\%$\uparrow$}) & 321 (\textbf{78.3\%$\uparrow$}) & 653 (\textbf{83.4\%$\uparrow$}) & 62 (\textbf{1.6\%$\uparrow$}) & 177 (\textbf{8.6\%$\uparrow$}) & 343 (\textbf{11.7\%$\uparrow$}) \\
\midrule
Random-BTTackler (\textbf{increase ratio})  & 110 (\textbf{74.6\%$\uparrow$}) & 322 (\textbf{73.1\%$\uparrow$}) & 634 (\textbf{72.8\%$\uparrow$}) & 101 (\textbf{62.9\%$\uparrow$}) & 305 (\textbf{69.4\%$\uparrow$}) & 621 (\textbf{74.4\%$\uparrow$}) & 97 (\textbf{59.0\%$\uparrow$}) & 259 (\textbf{58.9\%$\uparrow$}) & 486 (\textbf{58.3\%$\uparrow$})\\ 
\bottomrule
\end{tabular}
\label{efficiency}
\end{center}
\end{table*}

\begin{table}[htbp]
\caption{The time-saving of BTTackler in achieving best baseline accuracy on three tasks.}
\begin{center}
\begin{tabular}{lccc}
\toprule
\textbf{Task} & \textbf{Cifar10CNN}&\textbf{Cifar10LSTM}&\textbf{Ex96Trans}  \\
\midrule
SMAC            &4.64h&4.41h&4.51h  \\ \midrule
SMAC-BTTackler  &3.89h&2.31h&1.88h  \\ \midrule
\textbf{TSBA}   &\textbf{16\%} & \textbf{47\%} & \textbf{58\%}  \\
\bottomrule
\end{tabular}
\label{tsba}
\end{center}
\end{table}

In view of efficiency, we dive into why BTTackler can improve the accuracy of all tasks.
For example, we compare \textit{Random}, \textit{Random-ETR}, and \textit{Random-BTTackler} to illustrate the efficiency gains brought by BTTackler.
We recorded the number of finished trials in each task at 3 points, 1h, 3h, and 6h.
As we can see in Table \ref{efficiency}, both Random-BTTackler and Random-ETR ran more trials than Random within the same time budget.
Compared with Random-ETR, Random-BTTackler early terminated the bad trials according to the quality indicators instead of performance on the validation set.
For Cifar10CNN and Cifar10LSTM, BTTackler ran relatively fewer trials but gained better results, as Table \ref{effective} shows, which means less mistaking good trials for bad ones.
For the harder task, Ex96Trans, it is difficult for ETR to terminate trials for more promising chances, whereas BTTackler still worked well.
Taking the results in Figure \ref{fig-effective} and Table \ref{efficiency} together, Random-BTTackler transferred more energy into more promising trials. 

BTTackler's motivation is to reduce the HPO cost, and Time-Saving for Baseline Accuracy (TSBA) intuitively reflects that.
We measured TSBA on \textit{SMAC-BTTackler} in Table \ref{tsba}, which is $40.33\%$ on average, demonstrating that BTTackler can achieve the best accuracy of the baseline method in a smaller time cost.
\textit{SMAC-ETR} is not shown in Table \ref{tsba} because it performed worse than SMAC in most cases.
It is noteworthy that BTTackler brings higher efficiency in more complex models, like LSTM and Transformer.
As model complexity increases, more effort is needed to search for good hyperparameters, and bad trials may emerge more frequently.

\begin{figure}[htbp]
\centerline
{\includegraphics[width=3.1in]{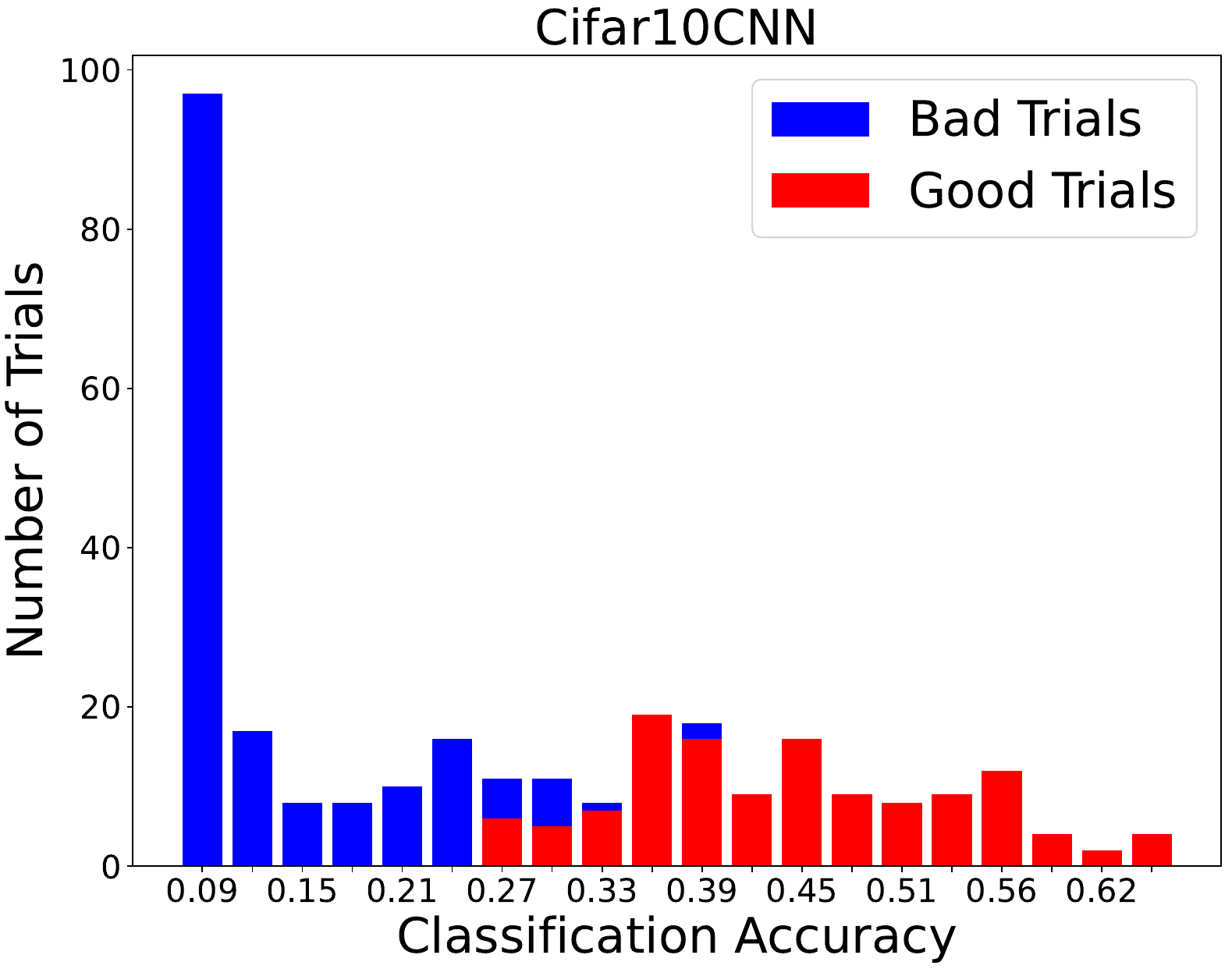}}
\caption{The performance distribution for both bad trials and good trials in the HPO task of Cifar10CNN.}
\label{motivation_single}
\end{figure}

We provide evidence of why quality indicators work. 
A performance distribution for both bad trials (alerted by quality indicators) and good trials (not alerted) in the HPO task of Cifar10CNN is presented in Figure \ref{motivation_single}.
A remarkable correlation exists between classification accuracy and quality indicators.
When bad trials indicated by quality indicators are largely terminated, a distribution transition may happen, increasing the probability of sampling hyperparameter configurations with higher accuracy.

The overhead of BTTackler is also a factor in the efficiency.
We recorded the operations timestamps of the processes in BTTackler for the three tasks.
The resource consumption of some quality indicators is large, such as the calculation of statistical values for high-dimensional gradient tensors. 
According to the timestamp records, if parallel computing is not used, the indicator calculation process will consume about 20\% more resources. 
If the main thread of model training is blocked, the time-saving effect brought by early stopping will be reduced. 
We use separate threads and workers to isolate the model training process, which mainly uses GPU resources, and the calculation process, which mainly consumes CPU resources in BTTackler.
Finally, the extra time overhead caused by BTTackler is limited to less than 5\% of the training time, which is negligible compared to the original HPO pipeline.


\subsection{Comparison of Quality indicators (RQ3)}
In this section, we use the simulator to calibrate quality indicators and analyze the individual impacts of the selected quality indicators in BTTackler.
Taking \textit{Random-BTTackler} as an example, we repeatedly replay all three HPO tasks by each unique quality indicator to illustrate the different impacts of the quality indicators, as shown in Table \ref{rules}.
The numbers denote how many bad trials each quality indicator claims in each task.
As the result shows, the quality indicators played different roles in the tasks.
In Cifar10CNN and Cifar10LSTM, the most significant quality indicators are \textit{ERG}, \textit{NMG}, and \textit{LAR}.
While in Ex96Trans, the most significant quality indicator is \textit{NMG}.
Some quality indicators did not show much functionality in one task but may perform well in others.

Notably, the complexity of HPO tasks leads to a distribution difference in the effectiveness of quality indicators.
The most effective quality indicators are concentrated on simpler HPO tasks, but more quality indicators are involved for harder HPO tasks.
Combining quality indicators leads to a maximum coverage of training problems, which should be the most effective strategy.
In practice, the quality indicators can be customized in BTTackler to adapt to different tasks.

\begin{table}[htbp]
\caption{Number of bad trials claimed by the quality indicators of BTTackler.}
\begin{center}
\begin{tabular}{m{1.8cm} m{0.55cm} m{0.55cm} m{0.55cm} m{0.55cm} m{0.55cm} m{0.55cm} m{0.55cm} }
\toprule
\textbf{Task} & \textbf{AGV}&\textbf{EAG}& \textbf{PLC}& \textbf{ERG}&\textbf{LAR}& \textbf{ULC}& \textbf{NMG} \\
\midrule
Cifar10CNN   & - & - & 24  & 317 & 62 & - & 93  \\ \midrule
Cifar10LSTM  & 8 & 20 & 1 & 296  & 105 & 19 & 134  \\ \midrule
Ex96Trans    & 1 & 3 & -   & 11   & 3 & 6 & 35  \\
\bottomrule
\end{tabular}
\label{rules}
\end{center}
\end{table}

\section{Conclusion} \label{conclusion}
In this paper, we presented BTTackler, a novel diagnosis-based HPO framework designed to identify training problems and terminate bad trials as soon as possible, thereby improving the efficiency of automated HPO.
We conducted a literature review of detecting DNN training problems to support the design of the quality indicators in BTTackler.
BTTackler traces every HPO trial and terminates it whenever quality indicators detect an obvious training problem, saving time and computational resources for further exploration of the hyperparameter space.
Our evaluation demonstrated that BTTackler outperformed the baselines, significantly improving efficiency and performance.

BTTackler brings a novel way to improve the HPO pipeline. 
Our future work will focus on some significant challenges for diagnosis-based HPO.
The first is the theoretical breakthrough in training diagnosis, where experimental expertise dominates the research right now. 
A promising research method is to analyze the distribution of weights and gradients during training. This should provide insights into both training diagnosis and HPO. 
The second is to work on quality indicator design methods to achieve greater gains and higher generalizability, which probably depends on the breakthrough of the training diagnosis theory.
The third is to find an approach to control the search space of HPO under a suitable and efficient scale, which may also benefit from training diagnosis.

\begin{acks}
This work was supported by the National Key Research and Development Program of China (2021YFB1715200), the National Natural Science Foundation of China (62021002, 92267203), and the National Engineering Research Center for Big Data Software.
\end{acks}

\bibliographystyle{ACM-Reference-Format}
\bibliography{sample-base}

\appendix

\section{Details of Literature Review} \label{literature}
This literature review focuses on the process of searching and screening detection methods for training problems for DNNs, including training diagnosis, debugging, and fault localization. 
The methodology consists of three main steps: (1) defining the search keywords and selecting the search engines, (2) collecting results and conducting Title-Abstract-Keywords (TAK) filtering, and (3) expanding the reference set using backward and forward snowballing.

To initiate the literature search, we utilized two major databases, i.e., Web of Science and Engineering Village. Then we applied the following search query to target DNN training problem detection indicators within the time range from Jan 2018 to June 2023. The search query was as follows: \textit{Title=("Deep learning" OR "Neural Network" OR "Training") AND Title=("Debugging" OR "Diagnosing" OR "Fault Localization" OR "Problem Detection")}. 
After the initial search, we got 245 and 189 articles from two databases, respectively. Then, we applied exclusion criteria to the initial set of articles to reserve articles related to training problem detection of DNNs. 
Specifically, we excluded articles on unrelated topics, including DNN-based software development tools, DNN-based applications, and post-training testing of DNNs. 
Some of the exclusion can be done with keywords, but more needs to be done manually. We closely examined the titles and abstracts of these articles to ensure their relevance to our topic. 
Subsequently, only six relevant articles were left for further search, as listed here: AutoTrainer\cite{zhang_autotrainer_2021}, Cockpit\cite{schneider2021cockpit}, DeepLocalize\cite{wardat_deeplocalize_2021}, ConvDiagnosis\cite{stamatescu2018diagnosing}, MODE\cite{ma_mode_2018}, and Umlaut\cite{schoop_umlaut_2021}. 
To further enrich the literature review, we employed backward and forward snowballing to explore other articles that referenced or were cited by the six articles left. 
Finally, we collected 14 articles to support our method, as shown in Table \ref{symptom}.

In the proposed framework, selecting appropriate quality indicators is crucial in ensuring the effectiveness and efficiency of achieving optimal experimental results. 
From the literature review, the previous research provided many methods, i.e., quality indicators, to quantify expertise in detecting training problems, as presented in Table \ref{symptom}.
As presented in Section \ref{background-training}, we need to select suitable quality indicators from the candidates according to the concerns and the motivation in HPO.
The five essential criteria that guide our selection are as follows:

\textbf{\textit{Non-Coding Error (NC).}}
In HPO cases, codes are fixed ahead of time, so the concerned training problems originate from bad hyperparameter configurations instead of coding errors.
So, the quality indicators of static coding errors are not considered. Such cases include inappropriate loss functions, missing or repeated activation functions, improper dropout rates, etc. 

\textbf{\textit{Non-Data Error (ND).}}
Similar to the above reason, the quality indicators of data errors are not in our scope. 
In the context of HPO for DNNs, data processing should be fixed among different trials. 
That is to say, any flaws in the training data or processing programs should be addressed before HPO.

\textbf{\textit{Deterministic instead of Descriptive (DD).}}
Some previous methods work on descriptions and suggestions for the training problems to construct a human-computer interaction work loop for developing DNNs.
We intend to make HPO more efficient with quality indicators, which requires the new HPO pipeline to be automated and unattended.
So, we adopt deterministic quality indicators that can be used directly in programs instead of descriptive ones.
While some descriptive quality indicators can be modified to be deterministic. 
We prioritize indicators that are with well-defined decision boundaries or rules. 
We carefully and selectively evaluate indicators that involve complex diagnostic processes.
Some methods are filtered out due to their reliance on learning methods or manual inspection for potential drawbacks, such as increased computational complexity, uncertainty, and reduced generalization ability.

\textbf{\textit{Acceptable Overhead (AO).}}
Quality indicators will be frequently called during the training of a large number of trials in HPO.
Thus, a notable overhead of a quality indicator may severely burden our proposed HPO pipeline.
The overhead includes various factors, such as computing complexity, memory usage, and IO costs. 
The quality indicators used in our framework must work under acceptable resource constraints.

\textbf{\textit{Acceptable Generalization (AG).}}
We also put the generalization ability of quality indicators into consideration.
The generalization ability includes compatibilities with mainstream deep learning libraries, popular DNN architectures, and commonly seen machine learning tasks.
To achieve acceptable generalization, we had to calibrate the quality indicators with the help of the simulator.
Maximizing performance on varied HPO tasks is feasible using the same quality indicators.
Therefore, we try to make the quality indicators conservative during training diagnosis, through which the generalization is maximized instead of the performance.

The above five criteria are used as guidance to pick up and modify the quality indicators from existing works.
The \textit{Used} column indicates whether the indicator is selected.
T, F, and P denote True, False, and Partially True.
The modification involves integrating similar indicators, transforming constants into customizable variables to improve adaptability, and replacing dependence on accuracy metrics with loss functions.
Besides, all the quality indicators are parallelized, as presented in Section \ref{implementation}. 

\begin{table*}[htbp]
\caption{Overview of Quality Indicators.
}
\begin{center}
\begin{tabular}{l|c|c|ccccc|c}
\toprule
\textbf{Reference} & \textbf{Indicator Name} & \textbf{Input}  
& \textbf{NC} 
& \textbf{ND} 
& \textbf{DD}
& \textbf{AO}
& \textbf{AG}
& \textbf{Used}
\\
\midrule

\multirow{4}{*}{AutoTrainer\cite{zhang_autotrainer_2021}}
& VanishingGradient & Gradient, Accuracy & T & T & T & T & P & \textbf{P} \\ 
\cline{2-9}
& ExplodingGradient & Gradient, Accuracy & T & T & T & T & T & \textbf{P}  \\ 
\cline{2-9}
& DyingReLU & Gradient, Accuracy & T & T & T & T & T & \textbf{T} \\ 
\cline{2-9}
& OscillatingLoss/SlowConvergence & Accuracy & T & T & T & T & P & \textbf{P} \\
\cline{2-9}
\hline

\multirow{3}{*}{CockPit\cite{schneider2021cockpit}}
& Incorrectly scaled data  & Data & T & F & T & F & T & \textbf{F}  \\
\cline{2-9}
& Vanishing gradients  & Gradient, Accuracy & T & T & T & T & P & \textbf{P} \\ 
\cline{2-9}
& Tuning learning rates  & Gradient, Accuracy & F & T & T & T & F & \textbf{F} \\
\hline

\multirow{2}{*}{DeepLocalize\cite{wardat_deeplocalize_2021}}
& Error with Activation/Function & Code, Model, Feature & F & T & T & T & F & \textbf{F} \\
\cline{2-9}
\cline{2-9}
& Error in Backaward & Model, Feature & T & T & T & T & P & \textbf{P} \\
\cline{2-9}
& Model does not Learn & Model, Feature & T & T & T & T & P & \textbf{P} \\
\hline

\multirow{1}{*}{ConvDiagnosis\cite{stamatescu2018diagnosing}}
& learning problems & Model, Weight, Gradient & T & T & F & T & P & \textbf{F} \\
\hline

\multirow{1}{*}{MODE\cite{ma_mode_2018}}
& Under/Over-fitting & Model, Data, Feature & T & T & F & P & T & \textbf{F} \\
\cline{2-9}
\hline

\multirow{8}{*}{Umlaut\cite{schoop_umlaut_2021}}
& Data Exceeds Limits & Data & T & F & P & F & P & \textbf{F} \\
\cline{2-9}
& NaN Loss & Loss & T & T & T & T & T & \textbf{T} \\
\cline{2-9}
& Incorrect Image Shape & Data & T & F & T & P & F & \textbf{F} \\
\cline{2-9}
& Unexpected Accuracy & Accuracy & T & T & T & T & P & \textbf{P} \\
\cline{2-9}
& Missing/Multiple Activation & Code, Model & F & T & T & T & P & \textbf{F} \\
\cline{2-9}
\cline{2-9}
& Abnormal Learning Rate & Code & F & T & T & T & F & \textbf{F} \\
\cline{2-9}
& Possible Overfitting & Loss & T & T & T & T & P & \textbf{P} \\
\cline{2-9}
& High Dropout Rate & Code & F & T & T & T & F & \textbf{F} \\
\hline

\multirow{8}{*}{DeepDiagnosis\cite{wardat_deepdiagnosis_2022}}
& ExplodingTensor & Loss, Weight, Gradient & T & T & T & T & T & \textbf{T} \\
\cline{2-9}
& UnchangeWeight & Weight & T & T & T & F & F & \textbf{F} \\
\cline{2-9}
& SaturatedActivation & Weight, Feature & T & T & T & T & P & \textbf{P} \\
\cline{2-9}
& DeadNode & Weight, Feature & T & T & T & T & T & \textbf{T} \\
\cline{2-9}
& OutofRange & Data, Feature & F & T & T & P & P & \textbf{F} \\
\cline{2-9}
& LossNotDecreasing & Loss & T & T & T & T & P & \textbf{P} \\
\cline{2-9}
& AccuracyNotIncreasing & Accuracy & T & T & T & T & P & \textbf{P} \\
\cline{2-9}
& VanishingGradient & Gradient & T & T & T & T & P & \textbf{P} \\
\hline

\multirow{6}{*}{DeepFD\cite{cao_deepfd_2022}}
& gap\_train\_test & Accuracy & T & T & T & T & P & \textbf{P} \\
\cline{2-9}
& test\_turn\_bad & Loss & T & T & T & T & P & \textbf{P} \\
\cline{2-9}
& slow\_converge & Accuracy & T & T & T & T & P & \textbf{P} \\
\cline{2-9}
& oscillating\_loss & Loss, Accuracy & T & T & T & T & P & \textbf{P} \\
\cline{2-9}
& dying\_relu & Gradient, Accuracy & T & T & T & T & P & \textbf{P} \\
\cline{2-9}
& gradient\_vanish/gradient\_explosion & Gradient, Accuracy & T & T & T & T & P & \textbf{P} \\
\cline{2-9}
\hline

\multirow{1}{*}{DeepXplore\cite{pei_deepxplore_2017}}
& neuron coverage & Gradient & T & T & T & T & T & \textbf{T} \\
\hline

\multirow{1}{*}{DeepFault\cite{eniser_deepfault_2019}}
& Suspicious Neurons & Gradient & T & T & F & P & T & \textbf{F} \\
\hline

\multirow{10}{*}{TFCheck\cite{braiek2019tfcheck}}
& Untrained Parameters & Weight & T & T & F & T & P & \textbf{F} \\
\cline{2-9}
& Poor Weight Initialization & Weight & T & T & T & T & P & \textbf{P} \\
\cline{2-9}
& Parameter Value Divergence & Weight & T & T & T & T & P & \textbf{P} \\
\cline{2-9}
& Parameter Unstable & Weight, Gradient & T & T & T & T & P & \textbf{P} \\
\cline{2-9}
& Activations Out of Range & Feature, Gradient & T & T & T & T & T & \textbf{T} \\
\cline{2-9}
& Neuron Saturation & Feature, Gradient & T & T & T & T & P & \textbf{P} \\
\cline{2-9}
& Dead ReLU & Feature, Gradient & T & T & T & T & P & \textbf{P} \\
\cline{2-9}
& Unable to fit & Data, Loss, Accuracy & T & P & T & P & F & \textbf{F} \\
\cline{2-9}
& Loss related issues & Loss & T & T & T & T & P & \textbf{P} \\
\cline{2-9}
& Unstable Gradients & Gradient & T & T & T & T & P & \textbf{P} \\
\hline

\multirow{3}{*}{Scram\cite{schoop2020scram}}
& Overfitting & Accuracy & T & T & T & T & P & \textbf{P} \\
\cline{2-9}
& Improper Data Normalization & Data & T & F & F & P & P & \textbf{F} \\
\cline{2-9}
& Unconventional Hyper-parameter & Code & F & T & F & T & F & \textbf{F} \\
\hline


\multirow{1}{*}{DetectingNN \cite{zhang2020detecting}}
& numerical bugs & Code, Model & F & T & T & P & P & \textbf{F} \\
\hline

\multirow{1}{*}{ReliabilityNN \cite{li2023reliability}}
& numerical defects & Code, Model & F & P & T & P & P & \textbf{F} \\

\bottomrule
\end{tabular}
\label{symptom}
\end{center}
\begin{flushleft}
\footnotesize
\textit{NC}, \textit{ND}, \textit{DD}, \textit{AO}, \textit{AG} denote Non-Coding Error, Non-Data Error, Deterministic, Accepted Overhead, Accepted Generalization. T, F, and P denote True, False, and Partially True.
\end{flushleft}
\end{table*}

Introducing some empirical parameters in the definitions of quality indicators is inevitable because the expertise in detecting training problems is not static or immutable.
To efficiently calibrate quality indicators in BTTackler, we propose a simulator-based method for evaluating them without conducting HPO trials.

\end{document}